\documentclass[10pt,twocolumn,letterpaper]{article}
\usepackage[accsupp]{axessibility}

\usepackage[pagenumbers]{cvpr} 
\usepackage{graphicx}
\usepackage{amsmath}
\usepackage{amssymb}
\usepackage{booktabs}
\usepackage{epsfig}
\usepackage{multirow}
\usepackage{makecell}
\usepackage{float}
\usepackage{subcaption}
\usepackage{diagbox}

\usepackage{colortbl}
\usepackage[dvipsnames]{xcolor}
\usepackage[linesnumbered,ruled,vlined]{algorithm2e}

\usepackage[dvipsnames]{xcolor}

\definecolor{cvprblue}{rgb}{0.21,0.49,0.74}
\usepackage[colorlinks=true,pagebackref,breaklinks,colorlinks,citecolor=cvprblue]{hyperref}

\newcommand{\blue}[1]{\textcolor{blue}{#1}}
\newcommand{\red}[1]{{\color{red}#1}}

\def\paperID{7045} 
\def\confName{CVPR}
\def\confYear{2024}

\title{Zero-Reference Low-Light Enhancement via Physical Quadruple Priors}

\author{Wenjing Wang\\
Peking University\\
\and
Huan Yang\\
01.AI\\
\and
Jianlong Fu\\
Microsoft Research Asia\\
\and
Jiaying Liu
\thanks{Corresponding author. This work was supported in part  by the National Natural Science Foundation of China under Grant 62332010, and in part by the Key Laboratory of Science, Technology and Standard in Press Industry (Key Laboratory of Intelligent Press Media Technology).}\\
Peking University\\
}

\begin{document}
\maketitle
\begin{abstract}
Understanding illumination and reducing the need for supervision pose a significant challenge in low-light enhancement. Current approaches are highly sensitive to data usage during training and illumination-specific hyper-parameters, limiting their ability to handle unseen scenarios.
In this paper, we propose a new zero-reference low-light enhancement framework trainable solely with normal light images. To accomplish this, we devise an illumination-invariant prior inspired by the theory of physical light transfer. This prior serves as the bridge between normal and low-light images.
Then, we develop a prior-to-image framework trained without low-light data.
During testing, this framework is able to restore our illumination-invariant prior back to images, automatically achieving low-light enhancement.
Within this framework, we leverage a pretrained generative diffusion model for model ability, introduce a bypass decoder to handle detail distortion, as well as offer a lightweight version for practicality.
Extensive experiments demonstrate our framework's superiority in various scenarios as well as good interpretability, robustness, and efficiency. Code is available on our \href{http://daooshee.github.io/QuadPrior-Website/}{project homepage}.
\end{abstract}    
\section{Introduction}
\label{sec:intro}

Restoring images in low-light conditions is an important and challenging task in computer vision.
The goal is to unveil concealed details in poorly lit areas, ultimately elevating the overall image quality. Over time, a large number of algorithms have been developed to address this challenge.
However, current methods exhibit limitations due to their dependence on supervisory information and their adaptability to unseen domains.
In the following, we review recent advancements and then present our primary contributions.

\vspace{1mm}
\noindent \textbf{Supervised Methods.}
Deep learning has significantly influenced the advancement of low-light enhancement.
In 2017, Li \etal~\cite{LLNet} introduced the initial deep-based low-light enhancement model using a straightforward auto-encoder.
Subsequently, a series of studies improved the network design by incorporating concepts from the Retinex theory~\cite{Enhance_RetinexNet,Enhance_KinD,URetinexNet}, Fourier transform~\cite{HuangLZYZHZX22}, image processing systems~\cite{HuangYHLD22}, semantics~\cite{SKF}, and adopting innovative architectures like Flow-based generative models~\cite{LLFlow}, vision transformers~\cite{Retinexformer} and diffusion models~\cite{PyDIff,DiffRetinex}.
In addition to RGB images, a body of research is dedicated to RAW data~\cite{Enhance_SID,SID_Motion,SMOID}, videos~\cite{StableLLVE}, and multi-modalities~\cite{Red_Flash,Flash}. 
While these models have achieved notable success, they rely on paired data for training, which can be inflexible and less robust in unforeseen scenarios. Recent studies explore how to reduce this dependency on supervision.

\begin{figure}[t]
    \centering
    \includegraphics[width=0.99\linewidth]{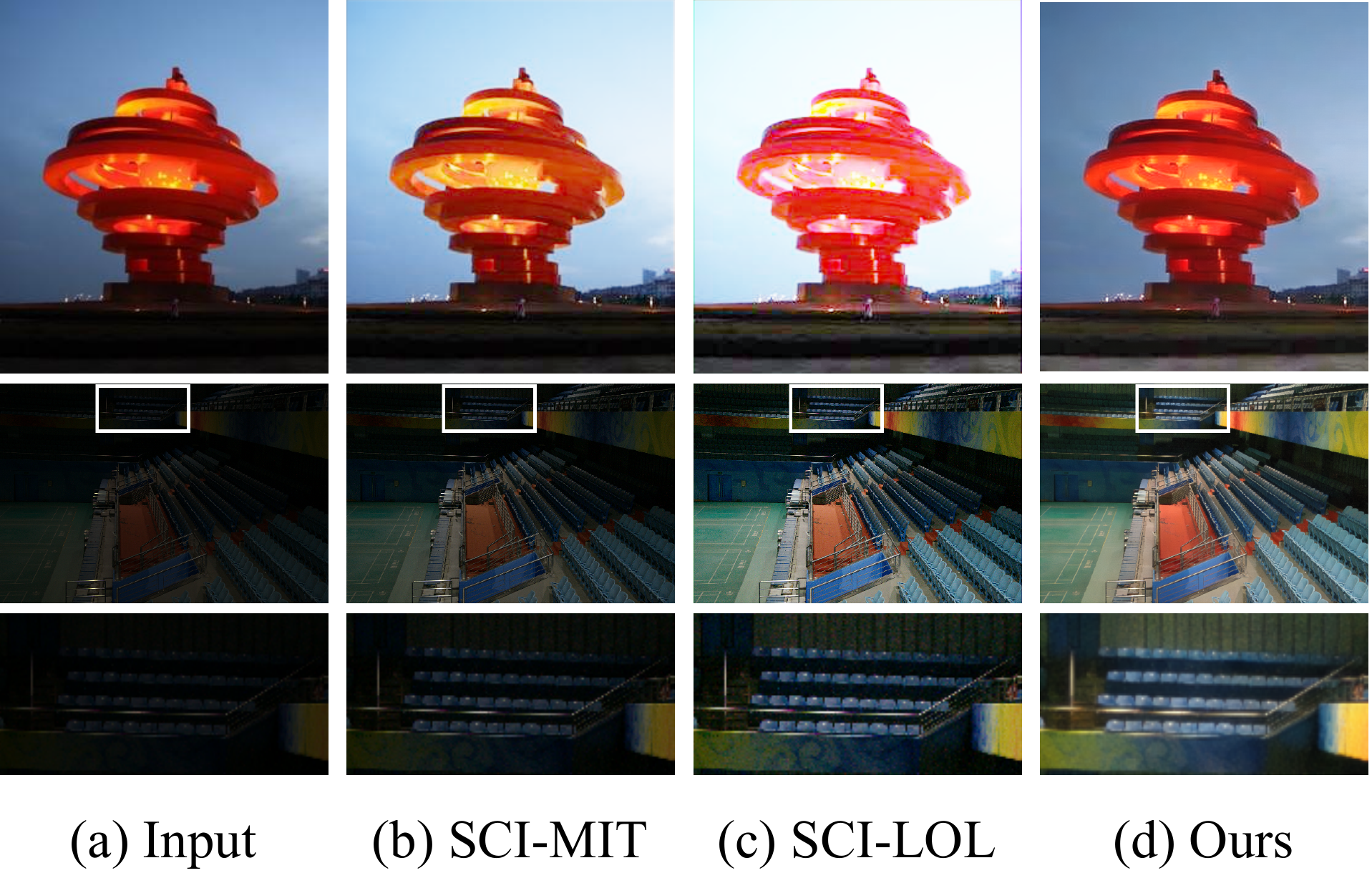}
    \caption{Comparison with a SOTA zero-reference method:
SCI~\cite{SCI}. The SCI model, trained on varied datasets like LOL~\cite{Enhance_RetinexNet} and MIT~\cite{FiveK}, yields diverse enhancement results. Nevertheless, none effectively maintains a consistent lighting effect across both dark and moderately dark images. In contrast, our model demonstrates greater robustness across various scenarios.}
    \label{fig:teaser}
    \vspace{-3mm}
\end{figure}

\noindent \textbf{Unsupervised Methods.}
Some work transitions from the strict requirement of pairwise pairings to only necessitating unpaired normal-low light data for training.
EnlightenGAN~\cite{Enhance_EnlightenGAN}, FlexiCurve~\cite{FlexiCurve}, and NeRCo~\cite{NeRCo} employ adversarial learning, where discriminators are constructed to guide low-light enhancement models, \ie, the generators.
CLIP-LIT~\cite{CLIP_LIT} learns prompts from images under varied illuminations.
PairLIE~\cite{PairLIE} instead learns adaptive priors from paired low-light instances of the same scene.
However, these methods retain specific training data requirements, limiting their ability to generalize to unknown scenarios.

\noindent \textbf{Zero-Reference Methods.}
``Zero-reference"~\cite{Enhance_ZeroDCE} refers to a special unsupervised setting where neither paired nor unpaired data is available for training.
This setting is more challenging, but offers greater flexibility in practical applications.
Traditional non-deep low-light enhancement algorithms~\cite{Enhance_HE,Enhance_LIME} can also be classified as zero-reference.
These methods primarily rely on manually designed strategies, such as histogram equalization~\cite{Enhance_HE} or Retinex decomposition~\cite{Enhance_LIME,LiLYSG18}.
As for deep models, Zero-DCE~\cite{Enhance_ZeroDCE} employs a neural network to predict the parameters of a pre-defined curve function and applies it to the input low-light image.
A suite of non-reference loss functions is employed to guide the enhancement process.
Subsequent works improved speed and curve forms~\cite{Enhance_ZeroDCE++,CuDi}.
As an alternative approach, RUAS~\cite{RUAS} introduces a neural architecture search strategy based on the Retinex rule.
It also implements several reference-free losses.
Furthermore, SCI~\cite{SCI} streamlines the iterative process in RUAS into a single step.

Despite the success of zero-reference methods, they often require careful parameter-tuning and can be sensitive to the distribution of training data.
For instance, in SCI~\cite{SCI}, varying training data leads to distinct enhanced appearances, causing over-exposure or under-exposure as shown in Fig.~\ref{fig:teaser}.
The crux of the matter is that these zero-reference models lack a genuine concept of illumination.
Learning lighting knowledge without reference and depending on artificially set parameters presents an challenging and unsolved problem.

\noindent \textbf{Our Contributions.}
In this paper, we propose a new zero-reference low-light enhancement framework to address the aforementioned challenges.
Our central idea is to develop an illumination-invariant prior and employ it as an intermediary between low-light and normal light images.
We devise a unique illumination prior, named the \textbf{physical quadruple prior}, originating from the Kubelka–Munk theory of light transfer.
Next, we construct a prior-to-image mapping framework \textbf{solely using typical normal-light images}, easily obtainable from the Internet or existing open-source visual datasets.
During this stage, the model learns the authentic concept of bright lighting from the image distribution.
Finally, when tested on low-light images, our physical quadruple prior automatically extracts illumination-invariant features, and prior-to-image mapping framework transfers these features to normal light images.
Throughout this process, low-light enhancement can be achieved without requiring any low-light data or illumination-relevant hyper-parameters.

The challenge lies in that, as our prior discards a considerable amount of illumination-relevant information, restoring it back to images is not a straightforward task.
To address this challenge, we capitalize on the exceptional capabilities of a pretrained large-scale generative model, Stable Diffusion (SD)~\cite{LDM}, and construct the prior-to-image mapping by integrating priors as conditions to control the SD model.
Unlike typical generative tasks that require high-quality data~\cite{DBLP:conf/cvpr/RuanMYH0FYJG23,DBLP:conf/mm/He00FY0CZ23,DBLP:conf/mm/Zhu0H0TCGSF23,DBLP:conf/mm/Zhu00HTYCGSFL23,DBLP:conf/mm/Ma00F022}, our framework is insensitive to data quality and is trained on one of the most readily available datasets: COCO~\cite{COCO}.
Since SD is originally designed for tasks other than restoration, it faces challenges in detail preservation.
Hence, we propose an bypass decoder to address the distortion issue, which also proves useful for other image processing tasks.
Finally, considering practical applications, our framework can be used to create a lighter zero-reference model on specific data.
By employing a CNN-transformer mixed model, we distill the complex multi-step optimization of a large diffusion model into a single forward pass within a lightweight network. 
This lightweight version maintains comparable performance while significantly improving inference speed and computational efficiency.

In summary, thanks to our physical quadruple prior, prior-to-image framework, the lightweight version, our approach combines \textit{interpretability}, \textit{robustness}, and \textit{efficiency}.
Experimental results demonstrate that our model attains favorable subjective and objective performance across diverse datasets.
Our main contributions are concluded as:
\begin{itemize}
    \item We present a zero-reference low-light enhancement model that utilizes an illumination-invariant prior as the intermediary between different illumination.
    Our model exhibits superior performance in various under-lit scenarios without relying on any specific low-light data.
    
    \item We establish the physical quadruple prior, a novel learnable illumination-invariant prior derived from a light transfer theory. This prior captures the essence of imaging under diverse lighting conditions, freeing low-light enhancement from dependence on reference samples or artificially set hyper-parameters.

    \item We develop an effective prior-to-image mapping system by incorporating the prior as a condition to control a pretrained large-scale generative diffusion model. We introduce a bypass decoder to address the distortion issue, and show that our model can be distilled into a lightweight version for practical application.

\end{itemize}
\section{Physical Prior-based Image Restoration}

\subsection{Motivation}

Developing illumination invariant features has a long history in the domain of image restoration.
One of the most representative invariants for low-light enhancement is the Retinex model~\cite{Retinex-original}.
This model posits that an image $X$ can be decomposed into illumination $I$ and illumination-invariant reflectance $R$, expressed as $X$~$=$~$I\odot R$, where $\odot$ denotes element-wise multiplication.
However, solving this decomposition is challenging.
Existing approaches either lean on human-crafted policies~\cite{Enhance_LIME} or rely on paired data with varying brightness levels~\cite{Enhance_RetinexNet,Enhance_KinD}, which lack robustness when faced with unknown scenarios.

\begin{figure}[t]
    \centering
    \includegraphics[width=0.99\linewidth]{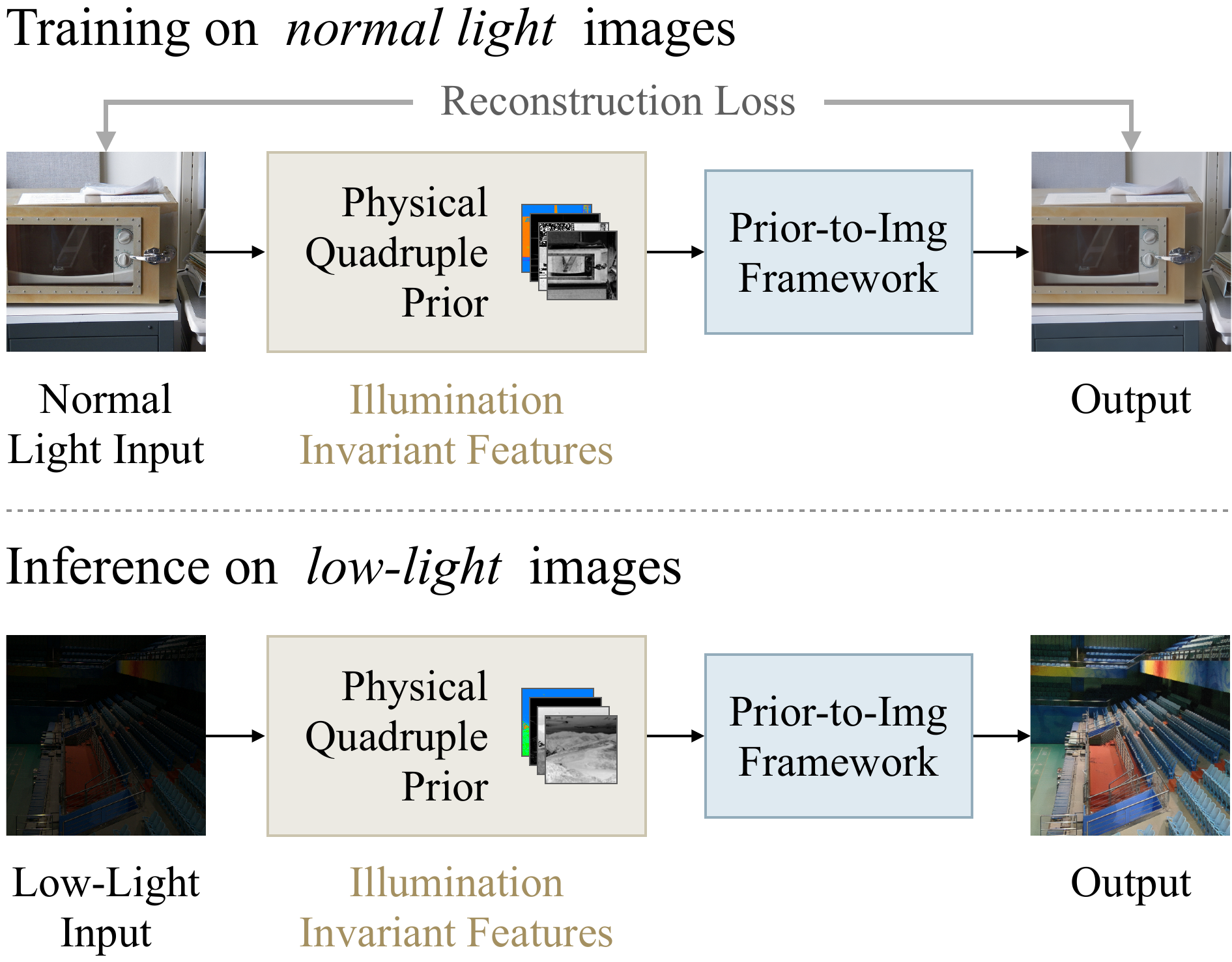}
    \caption{The overall methodology of our zero-reference low-light enhancement approach.
    Our model is trained to reconstruct images from an illumination-invariant prior (the physical quadruple prior) in the normal light domain.
    During testing, the model extracts illumination-invariant priors from low-light images and reconstructs them into normal light images.}
    \label{fig:overall}
    \vspace{-3mm}
\end{figure}

\begin{figure*}[t]
    \centering
    \includegraphics[width=0.99\linewidth]{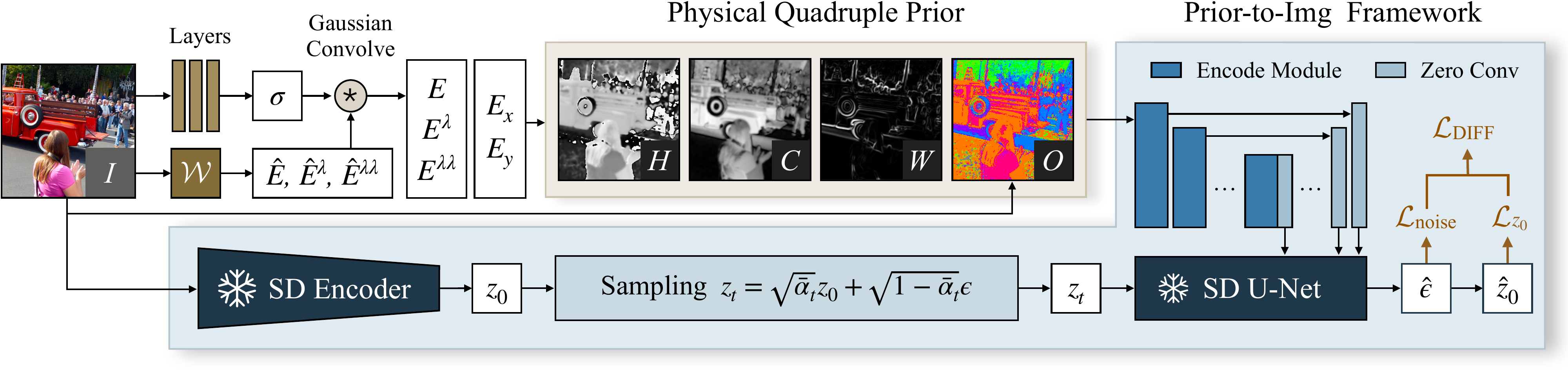}
    \caption{
    Our illumination-invariant prior and the training process for our prior-to-image model framework.
    We start by predicting the physical quadruple prior from the input image $I$.
    During the training phase, the model dynamically learns the linear mapping $\mathcal{W}$ and the layers for predicting the scale $\sigma$.
    In the process of reconstructing priors into images, a static SD encoder extracts the latent representation $z_0$ from the input image $I$.
    Following this, we sample noisy latent $z_t$ based on $z_0$.
    Finally, the physical quadruple prior is encoded by convolutional and transformer modules, and is then merged with a frozen SD U-net to predict both noise $\epsilon$ and $z_0$.}
    \label{fig:framework}
\end{figure*}

Instead of decomposing images into illumination-invariant and illumination-related information, we suggest to solely extract illumination-invariant features from the input image and subsequently generate illumination-related information to reconstruct the image.
In contrast to traditional Retinex-based deep models~\cite{Enhance_RetinexNet,Enhance_KinD}, our framework ensures that our deep model can comprehend lighting without the need for paired data.

The overall training and inference pipeline of our method is illustrated in Fig.~\ref{fig:overall}.
We introduce a physical quadruple prior derived from the Kubelka-Munk theory~\cite{gevers2012color} to extract illumination-invariant features.
While training on normal light images, the model simultaneously learns how to reconstruct images from priors and learns the illumination distribution of natural images.
Acknowledging the complexity of simultaneously addressing these two tasks, we make use of a pre-trained diffusion generative model.
During testing, low-light images are initially mapped to the physical quadruple prior and subsequently reconstructed.
Given the illumination-invariant nature of our prior, it extracts comparable features from low-light images as it does from normal light images.
Combined with the fact that our prior-to-image model is specifically trained to reconstruct priors into normal light images, our model achieves low-light enhancement without the need for low-light data.
In the following, we elaborate on the details of each component.

\subsection{Learnable Illumination-Invariant Prior}
\label{sec:prior}

\noindent \textbf{Physical Quadruple Priors.}
We start from the Kubelka-Munk theory~\cite{gevers2012color} of light transfer.
Given wavelength $\lambda$, the energy of the incoming spectrum at spatial location $\mathbf{x}$ on the image plane is modeled as
\begin{align}
\label{eq:kubelka_munk}
 E(\lambda,\mathbf{x}) = e(\lambda, \mathbf{x})\left((1-i(\mathbf{x}))^2R_\infty(\lambda,\mathbf{x})+i(\mathbf{x})\right),
\end{align}
where $e(\lambda, \mathbf{x})$ denotes the spectrum of the light source, $i(\mathbf{x})$ the specular reflection, and $R_\infty(\lambda,\mathbf{x})$ the material reflectivity.
Note that when the object is matte, \ie, $i(\mathbf{x})\approx0$, Eq.\,(\ref{eq:kubelka_munk}) can be reduced to
\begin{align}
 E(\lambda,\mathbf{x}) = e(\lambda, \mathbf{x})R_\infty(\lambda,\mathbf{x}),
\end{align}
which is the same as the Retinex model.
It means that the Retinex theory is a special case of Eq.\,(\ref{eq:kubelka_munk}).

First of all, we denote some variables for simplicity
\begin{align}
    E^{\lambda} &= \frac {\partial E(\lambda,\mathbf{x})} {\partial \lambda}, \;
    R_\infty^{\lambda} = \frac {\partial R_\infty(\lambda,\mathbf{x})} {\partial \lambda},
    \\
    E^{\lambda\lambda} &= \frac {\partial^2 E(\lambda,\mathbf{x})} {\partial \lambda^2}, \;
    R_\infty^{\lambda\lambda} = \frac {\partial^2 R_\infty(\lambda,\mathbf{x})} {\partial \lambda^2}.
\end{align}
Intuitively, $E$ represents spectral intensity, $E^{\lambda}$ signifies spectral slope, and $E^{\lambda\lambda}$ denotes spectral curvature. 

Following \cite{GeusebroekBSG01}, through simplifying assumptions, we can obtain a series of invariants from Eq.\,(\ref{eq:kubelka_munk}).
The primary idea is to eliminate $i$ and $e$, retaining solely $R_\infty$.
As $R_\infty$ is about material property and is independent of illumination, the derived variable will exhibit illumination invariance.

\begin{itemize}
    \item Assuming \textit{equal energy} illumination, \ie, $e(\lambda, \mathbf{x})$ is reduced to $\lambda$-independent $\tilde{e}(\mathbf{x})$, and Eq.\,(\ref{eq:kubelka_munk}) is reduced to
\begin{align}
\label{eq:equal_illumination}
 E(\lambda,\mathbf{x}) = \tilde{e}(\mathbf{x})\left((1-i(\mathbf{x}))^2R_\infty(\lambda,\mathbf{x})+i(\mathbf{x})\right),
\end{align}
Substituting Eq.\,(\ref{eq:equal_illumination}) into $E^{\lambda}/E^{\lambda\lambda}$ gives
\begin{align}
\frac{E^{\lambda}} {E^{\lambda\lambda}} = \frac {\tilde{e}(\mathbf{x})(1-i(\mathbf{x}))^2R_\infty^{\lambda}} {\tilde{e}(\mathbf{x})(1-i(\mathbf{x}))^2R_\infty^{\lambda\lambda}} = \frac{R_\infty^{\lambda}} {R_\infty^{\lambda\lambda}},
\end{align}
where illumination properties $i$ and $e$ are eliminated. As the material property $R_\infty$ is independent of illumination, it establishes the illumination-invariance of $E^{\lambda}/E^{\lambda\lambda}$.
Now we derive our first illumination invariant,
\begin{align}
\label{eq:H}
     H = \arctan \left(  {E^{\lambda}} / {E^{\lambda\lambda}} \right).
\end{align}
    \item \textit{Further} assuming that the surface is \textit{matte}, \ie $i(\mathbf{x})\approx0$, then Eq.\,(\ref{eq:kubelka_munk}) is reduced to
\begin{align}
\label{eq:matte}
 E(\lambda,\mathbf{x}) = \tilde{e}(\mathbf{x})R_\infty(\lambda,\mathbf{x}),
\end{align}
    Similarly, we derive another illumination invariant,
\begin{align}
\label{eq:C}
    C & = \log \left( \frac { (E^{\lambda})^2 + (E^{\lambda\lambda})^2} {E(\lambda,\mathbf{x})^2} \right) \nonumber \\
    & = \log \left( \frac { (R_\infty^{\lambda})^2 + (R_\infty^{\lambda\lambda})^2} {R_\infty(\lambda,\mathbf{x})^2} \right).
\end{align}
    \item \textit{Further} assuming \textit{uniform} illumination,  \ie, $\tilde{e}(\mathbf{x})$ is reduced to a parameter $\bar{e}$, and Eq.\,(\ref{eq:kubelka_munk}) is reduced to
\begin{align}
\label{eq:uniform}
 E(\lambda,\mathbf{x}) = \bar{e}R_\infty(\lambda,\mathbf{x}),
\end{align}
    Similarly, we derive our third illumination invariant,
\begin{align}
\label{eq:W}
    W &= \tan \left( \bigg| \frac {\partial E(\lambda,\mathbf{x})} {\partial \mathbf{x}} \frac 1 {E(\lambda,\mathbf{x})} \bigg| \right) \nonumber \\
    &= \tan \left( \bigg| \frac {\partial R_\infty(\lambda,\mathbf{x})} {\partial \mathbf{x}} \frac 1 {R_\infty(\lambda,\mathbf{x})} \bigg| \right).
\end{align}
\end{itemize}

The Kubelka-Munk theory~\cite{gevers2012color} is effective for grayscale images but falls short in describing colors. The three aforementioned illumination invariants loss part of the color information, so we add some additional color information.
We adopted a straightforward one: the relative relationship between pixel values of the three RGB channels.

\begin{itemize}
    \item Assuming that illumination maintains the order of colors, we propose the order of the RGB three channels as a fundamental illumination-invariant feature, denoted as $O$.
\end{itemize}

\vspace{1mm}
\noindent \textbf{Learning through Neural Networks.}
We follow Gaussian color models~\cite{gevers2012color} and CIConv~\cite{CIConv} to obtain priors from RGB images. First, we estimate the observed energy $\hat{E}$ along with its derivatives $\hat{E}^\lambda$ and $\hat{E}^{\lambda\lambda}$ via linear mapping: 
\begin{align}
\label{eq.rgb_gaussian}
  \begin{bmatrix}
  \hat{E}(x,y)\\
  \hat{E}^\lambda(x,y)\\
  \hat{E}^{\lambda\lambda}(x,y)\\
  \end{bmatrix}
  \!=\!
  \mathcal{W}
  \begin{bmatrix}
  R(x,y)\\G(x,y)\\B(x,y)
  \end{bmatrix},
\end{align}
where $x$ and $y$ denote positions in the image, and $\mathcal{W}$ is a $3\times3$ matrix.
In \cite{gevers2012color,CIConv}, $\mathcal{W}$ is manually defined. 
We instead learn it from the distribution of natural images through our prior-to-image framework.

The spatial derivative $\partial E / \partial \mathbf{x}$ in Eq.\,(\ref{eq:W}) is computed in both the x- and y-direction, denoted as $\partial E / \partial \mathbf{x}=(E_x, E_y)$, with its magnitude given by $|\partial E / \partial \mathbf{x}| = \sqrt{E_x^2 + E_y^2}$.
Finally, $E$, $E_x$, and $E_y$ are estimated by convolving $\hat{E}$ with Gaussian color smoothing and derivative filters of scale $\sigma$. 
$\sigma$ is predicted from the input image.
Similarly, $E^{\lambda}$ is obtained from $\hat{E}^{\lambda}$, and $E^{\lambda\lambda}$ is obtained from $\hat{E}^{\lambda\lambda}$.
Now we can compute $H$, $C$, and $W$ from the input image.

Our last illumination invariant, the order of RGB channels, is defined as three channels as follows,
\begin{align}
    O(x,y) = \left[ O_R(x,y), O_G(x,y), O_B(x,y) \right], 
\end{align}
where $O_R$ represents the order of the $R$ channel in RGB, normalized to [-1,1]. $O_G$ and $O_B$ are treated similarly.

Finally, $H$, $C$, $W$, and $O$ are concatenated in the channel dimension to form our physical quadruple prior.

\vspace{1mm}
\noindent \textbf{Physical Explanation.}
Firstly, the mathematical form indicates that $W$ represents the intensity-normalized spatial derivatives of the spectral intensity.
As for $H$, according to \cite{gevers2012color}, it is associated with the hue, \ie, $\arctan(\lambda_\text{max})$ of the material.
As for $C$, within a color circle based on spectral wavelengths, hue represents the angle while chroma is the distance from the center.
Additionally, when converting a Cartesian coordinate $(a, b)$ to polar, the angle becomes $\arctan(b/a)$, and the radius becomes $\sqrt{(a)^2+(b)^2}$.
Connecting this with Eq.\,(\ref{eq:H}) and Eq.\,(\ref{eq:C}), we find $C$ associates with chroma.
A visualization can be found in Fig.~\ref{fig:framework}. More analysis and comparisons will be presented in Sec.~\ref{sec:ablation}.

\subsection{Prior-to-Image via Diffusion Models}

Ideally, we want to retain all illumination-invariant information while discarding lighting-relevant information.
However, achieving this decomposition is challenging and remains a unsolved problem in image modeling.
Despite our physical quadruple prior, \ie, $H$, $C$, $W$, and $O$, capturing illumination-independent information from various perspectives, some information is still lost.
Consequently, reconstructing images from the prior is non-trivial.

\begin{figure}
    \centering
    \includegraphics[width=0.99\linewidth]{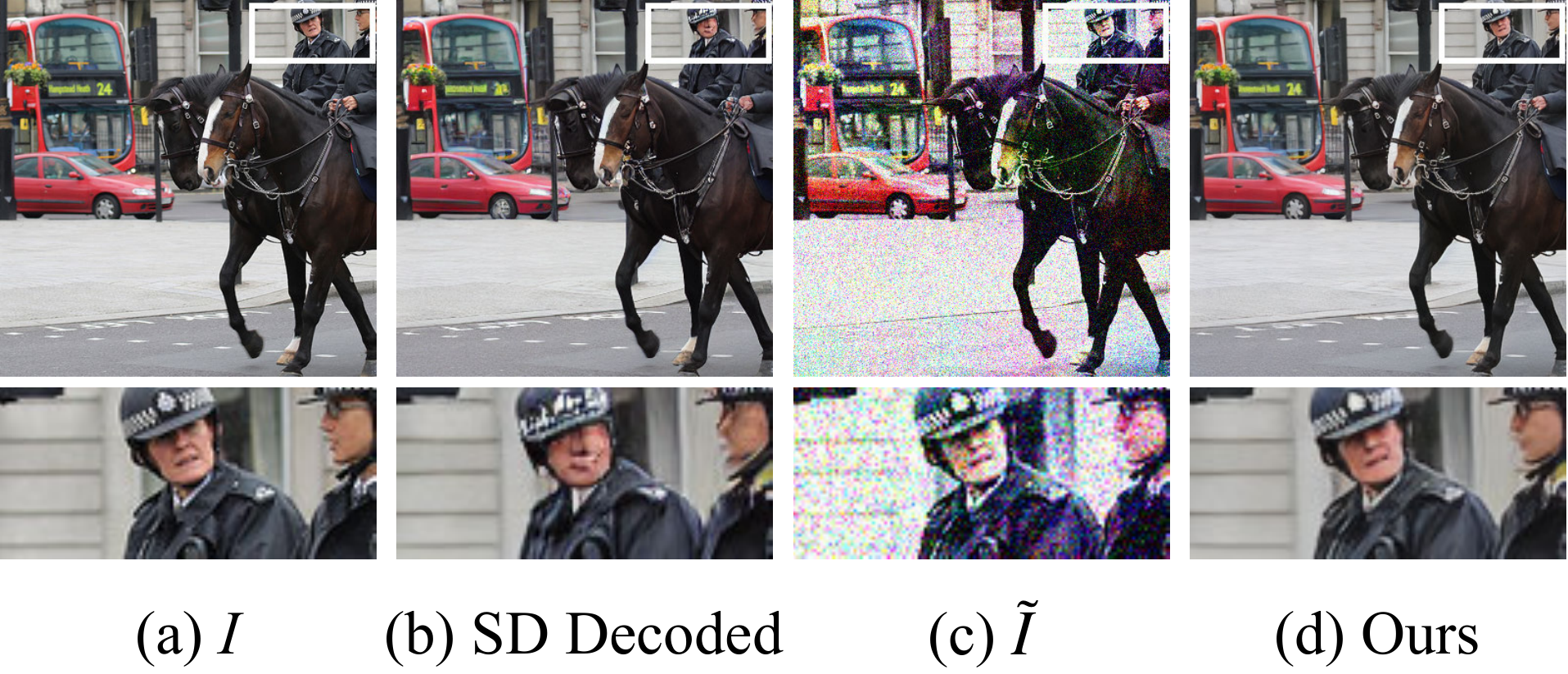}
    \vspace{-1mm}
    \caption{Image restoration effect of the SD decoder and ours. (a) Input image $I$, from which we extract latent $z_0$. (b) $z_0$ decoded by the SD decoder. (c) The distorted version of $I$. (d) $z_0$ decoded by our decoder using the encoder features from $\tilde{I}$.
    }
    \label{fig:ae_train_compare}
\end{figure}

Instead of focusing on improving the illumination-invariant prior, we propose leveraging the capabilities of a large-scale generative model to directly complete the missing information.
We employ Stable Diffusion v1-5~\cite{LDM} and convert it into a conditional generative mode using the ControlNet~\cite{ControlNet} framework.
Our physical quadruple prior serves as the condition to control the SD model.

The overall framework is illustrated in Fig.~\ref{fig:framework}.
During training, a frozen SD encoder is employed to map the image $I$ into a compressed latent representation $z_0$.
We then sample $z_t$ at a random time step $t\in\{1,...,T\}$ using
\begin{align}
\label{eq:z_t}
    z_{t}=\sqrt{\bar{\alpha}_t}z_0+\sqrt{1-\bar{\alpha}_t} \epsilon,
\end{align}
where $\{\bar{\alpha}_t\}$ is a sequence of pre-defined parameters~\cite{DDPM}.
The training objective is to predict $\epsilon$ from $z_{t}$ and our prior. Originally, SD utilizes a U-Net to predict $\epsilon$ from $z_{t}$, and this U-Net is now frozen.
A set of encode modules are added to extract features from our quadruple prior.
These features are then incorporated into the SD U-Net.
The zero convolution strategy~\cite{ControlNet} is adopted to ensure that, at the beginning of training, new layers do not influence the original SD.
During testing, given an input image $I$, we extract the physical quadruple prior and use it as the condition to predict $z_0$ in the reverse diffusion process.
Subsequently, $z_0$ is projected back to the image space through a decoder.

\begin{figure}
    \centering
    \includegraphics[width=0.99\linewidth]{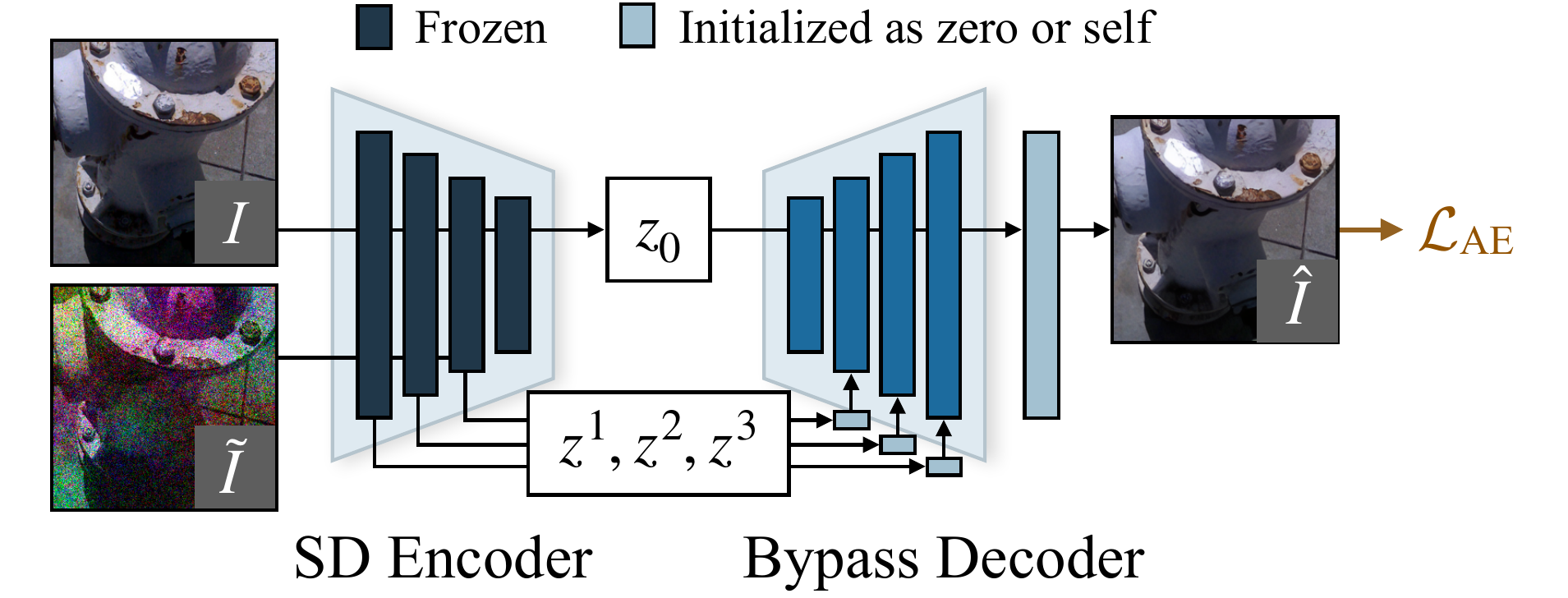}
    \caption{The training strategy of our bypass decoder. We distort the input image $I$ into $\tilde{I}$, and allow the decoder to reconstruct $I$ using encoder features from the distorted $\tilde{I}$.}
    \label{fig:autoencoder}
\end{figure}

\begin{figure}
    \centering
    \includegraphics[width=0.99\linewidth]{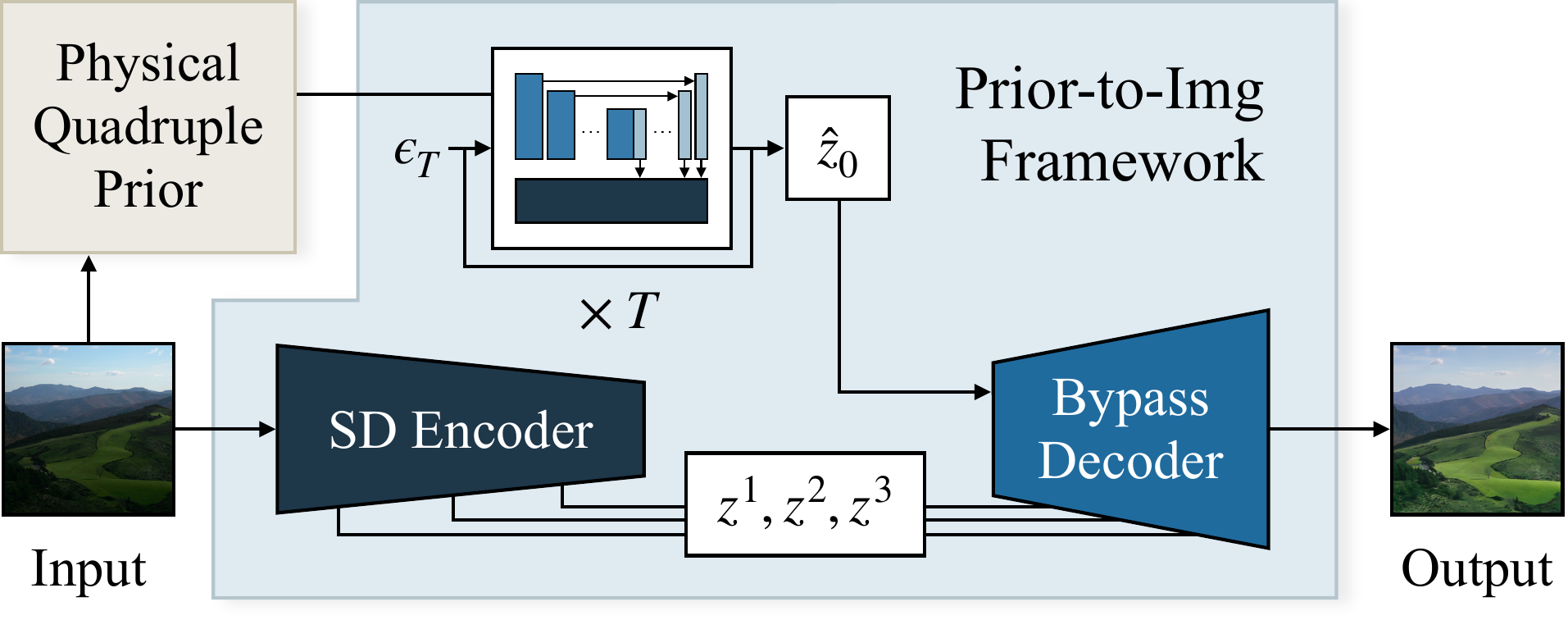}
    \vspace{-1mm}
    \caption{The inference pipeline of our overall framework. Given a low-light image, we extract its physical quadruple prior. Then, this prior serves as the condition for predicting the latent representation $\hat{z}_0$ from pure noise $\epsilon_T$.
    Lastly, the bypass decoder utilizes features extracted by the encoder from the low-light image to map the predicted $\hat{z}_0$ back into images.
    }
    \label{fig:inference}
\end{figure}

While ControlNet has demonstrated success in various applications, applying it directly suffers from issues including \textit{slow convergence}, \textit{detail degradation}, and \textit{dependence on text prompts}.
To address these challenges and make it more suitable for our image restoration task, we implement the following improvements.
\begin{itemize}
    \item In typical diffusion models, the training objective is to predict the Gaussian noise term:
\begin{align}
    \mathcal{L}_{\text{noise}} = ||\epsilon-\hat{\epsilon}||^2_2. 
\end{align}
    We additionally minimize the difference in terms of $z_0$, which can accelerate convergence. Combining Eq.\;(\ref{eq:z_t}), we derive
\begin{align}
\label{eq:z0}
    \mathcal{L}_{z_0} = ||z_0-\hat{z}_0||^2_2 = 
    ||z_0 - \frac{z_t-\sqrt{1-\bar{\alpha}_t}\hat{\epsilon}} {\sqrt{\bar{\alpha}_t}} ||^2_2.
\end{align}
    We simple combine these two losses as the final objective
\begin{align}
    \mathcal{L}_{\text{DIFF}} = \mathcal{L}_{z_0} + \mathcal{L}_{\text{noise}}.
\end{align}
    \item SD employs an auto-encoder (AE) to compress the image $I$ into a latent representation $z_0$, reducing computational cost. However, the auto-encoder introduces severe detail distortion.
    As shown in Fig.~\ref{fig:ae_train_compare}(b), the face of the mounted police is completely distorted.
    To mitigate this issue, we support the decoder with features from the encoder and devise an effective fine-tuning strategy.
    As shown in Fig.~\ref{fig:autoencoder}, during training, we distort the input image $I$ with random illumination jittering and noise, resulting in $\tilde{I}$.
    The decoder then restores $z_0$, combining the features $z^1$, $z^2$, $z^3$ extracted from $\tilde{I}$. 
    This progress guides the decoder to capture details from $\tilde{I}$ while preserving the illumination characteristics of $I$.
    We introduce several convolutional layers for feature fusion and a residual block for post-processing. These additional layers are initialized as zero or self, ensuring they have minimal impact on the original decoding process at the beginning of training.
    The new decoder is named bypass decoder.
    As shown in Fig.~\ref{fig:ae_train_compare}(d), noticeable detail restoration is achieved through our bypass decoder.
    During testing, features from the input image assist in the latent decoding process, as illustrated in Fig.~\ref{fig:inference}. Our bypass decoder utilizes $z^1$, $z^2$, and $z^3$, extracted from the input image, to reconstruct details and maintain the enhanced illumination within $\hat{z}_0$.
    \vspace{1mm}
    \item Stable Diffusion is originally designed as a text-to-image model. However, requiring users to provide text for low-light enhancement is inconvenient. To address this, we set the text input to always be an empty string.
\end{itemize}

\vspace{1mm}
\noindent \textbf{Denoising.}
Noise poses a significant challenge in low-light enhancement.
Although our prior isn't designed for denoising purposes, we implement a simple strategy to suppress noise.
During training, we apply random Gaussian-Poisson compound noise to the input image $I$ while extracting the physical quadruple prior.
This approach guides the model to disregard high-frequency details and concentrate solely on low-frequency illumination-invariant information.

\vspace{1mm}
\noindent \textbf{Distillation for Efficiency.}
Diffusion models require multi-step optimization in inference. Even with DPM-Solver++~\cite{DPM_solver_plusplus}, 10 steps are still cumbersome.
In the pursuit of practicality, our framework can create a more lightweight version.
In short, we construct a lightweight U-net consists of residual blocks and integrate transformer blocks from Restormer~\cite{Restormer} at the bottleneck.
Transformers have proven to be effective in low-level vision~\cite{TTSR,10239462,DBLP:conf/eccv/QiuYFF22,DBLP:conf/cvpr/Liu0FQ22}.
Then, we make 1.7k samples using our framework to teach the lightweight model.
The training objective is solely the L1 loss.
More details are provided in the supplementary.
\begin{table*}[t]
    \centering
    \caption{Benchmarking results for low-light enhancement.
    Among unsupervised methods, we highlight the top-ranking scores in \red{red} and the second in \blue{blue}. Additionally, we denote the training set used by each model. ``LOL+" indicates a fusion of LOL and other datasets.}
    \vspace{-2mm}
    \scriptsize
    \begin{tabular}{l|l|c|cccc|cccc|cc}
        \toprule
        \multicolumn{2}{l|}{Datasets} & Train Set & \multicolumn{4}{c|}{LOL~\cite{Enhance_RetinexNet,Enhance_DRBN}} & \multicolumn{4}{c|}{MIT-Adobe FiveK~\cite{FiveK}} & \multicolumn{2}{c}{Unpaired Sets} \\
        \midrule
        \multicolumn{2}{l|}{Metrics} & & PSNR$\uparrow$ & SSIM$\uparrow$ & LPIPS$\downarrow$ & LOE$\downarrow$ & PSNR$\uparrow$ & SSIM$\uparrow$ & LPIPS$\downarrow$ & LOE$\downarrow$ & BRISQUE$\downarrow$ & NL$\downarrow$ \\
        \midrule
        \multirow{7}{*}{Supervised}
        & Retinex-Net~\cite{Enhance_RetinexNet} & LOL & 16.19 & 0.403 & 0.534 & 0.346 & 12.30 & 0.687 & 0.258 & 0.244 & 27.10 & 3.254 \\
        & KinD~\cite{Enhance_KinD} & LOL & 20.21 & 0.814 & 0.147 & 0.245 & 14.71 & 0.756 & 0.176 & 0.174 & 26.89 & 0.700 \\
        & KinD++~\cite{Kind++} & LOL & 16.64 & 0.662 & 0.410 & 0.288 & 15.76 & 0.650 & 0.319 & 0.176 & 26.16 & 0.431 \\
        & URetinex-Net~\cite{URetinexNet} & LOL & 20.93 & 0.854 & 0.104 & 0.245 & 14.10 & 0.734 & 0.182 & 0.187 & 23.80 & 1.319 \\
        & Retinexformer~\cite{Retinexformer} & LOL &  28.48 & 0.877  & 0.117 & 0.256 & 13.87 & 0.692 & 0.222 & 0.224 & 14.77 & 1.064 \\
        & Retinexformer~\cite{Retinexformer} & MIT &  13.02 & 0.426 & 0.365 & 0.280 & 24.93 & 0.907 & 0.063 & 0.162 & 24.13 & 0.684 \\
        & DiffLL~\cite{Diffll} & LOL+ & 28.54 & 0.870 & 0.102 & 0.253 & 15.81 & 0.719 & 0.244 & 0.213 & 14.96 & 0.888 \\
        \midrule 
        \multirow{16}{*}{Unsupervised}
        & ExCNet~\cite{ExCNet} & test images & 16.29 & 0.455 & 0.380 & 0.295 & 14.21 & 0.719 & 0.197 & 0.197 & 19.03 & 1.563 \\
        & EnlightenGAN~\cite{Enhance_EnlightenGAN} & own data & 18.57 & 0.700 & 0.302 & 0.291 & 13.28 & 0.738 & 0.203 & 0.199 & 20.65 & 0.779 \\
        & PairLIE~\cite{PairLIE} & LOL+ & \blue{19.70} & \blue{0.774} & \blue{0.235} & \red{0.278} & 10.55 & 0.642 & 0.273 & 0.225 & 29.84 & 1.471 \\
        & NeRCo~\cite{NeRCo} & LSRW~\cite{LSRW} & 19.67 & 0.720 & 0.266 & 0.310 & \blue{17.33} & 0.767 & 0.208 & 0.213 & 22.81 & \blue{0.603} \\
        & CLIP-LIT~\cite{CLIP_LIT} & own data & 14.82 & 0.524 & 0.371 & 0.320 & 17.00 & 0.781 & \blue{0.159} & 0.194 & 23.44 & 1.962 \\ 
        & ZeroDCE~\cite{Enhance_ZeroDCE} & own data & 17.64 & 0.572 & 0.316 & 0.296 & 13.53 & 0.725 & 0.201 & 0.191 & 21.76 & 1.569 \\
        & ZeroDCE++~\cite{Enhance_ZeroDCE++} & own data & 17.03 & 0.445 & 0.314 & 0.391 & 12.33 & 0.408 & 0.280 & 0.417 & 19.34 & 1.150 \\
        & RUAS~\cite{RUAS} & MIT & 13.62 & 0.462 & 0.346 & 0.292 & 9.53 & 0.610 & 0.301 & 0.272 & 29.91 &  2.091 \\
        & RUAS~\cite{RUAS} & LOL & 15.47  & 0.490 & 0.305 & 0.330 & 5.15 & 0.373 & 0.669 & 0.399 & 44.70 &  3.312 \\
        & RUAS~\cite{RUAS} & FACE~\cite{DARKFACE} & 15.05 & 0.456 & 0.371 & 0.292 & 5.00 & 0.366 & 0.685 & 0.398 & 46.21 & 3.633 \\
        & SCI~\cite{SCI} & MIT & 11.67 & 0.395 & 0.361 & 0.286 & 16.29 & \red{0.795} & \red{0.143} & \red{0.165} & \blue{16.73} & 0.853 \\
        & SCI~\cite{SCI} & LOL+ & 16.97 & 0.532 & 0.312 & 0.289 & 7.83 & 0.573 & 0.360 & \blue{0.187} &  24.46 &  1.893\\
        & SCI~\cite{SCI} & FACE~\cite{DARKFACE} & 16.80 & 0.543 & 0.322 & 0.297 & 10.95 & 0.684 & 0.272 & 0.205 & 18.33 & 1.335 \\
        \cmidrule{2-13}
        & \textbf{Ours} & COCO~\cite{COCO} & \red{20.31} & \red{0.808} & \red{0.202} & \blue{0.281} & \red{18.51} & \blue{0.785} & 0.163 & 0.188 & \red{14.64} & \red{0.423}  \\
        \bottomrule
    \end{tabular}
    \label{tab:lol_fivek_unpaired}
\end{table*}

\begin{figure*}[t]
    \centering
    \includegraphics[width=0.99\linewidth]{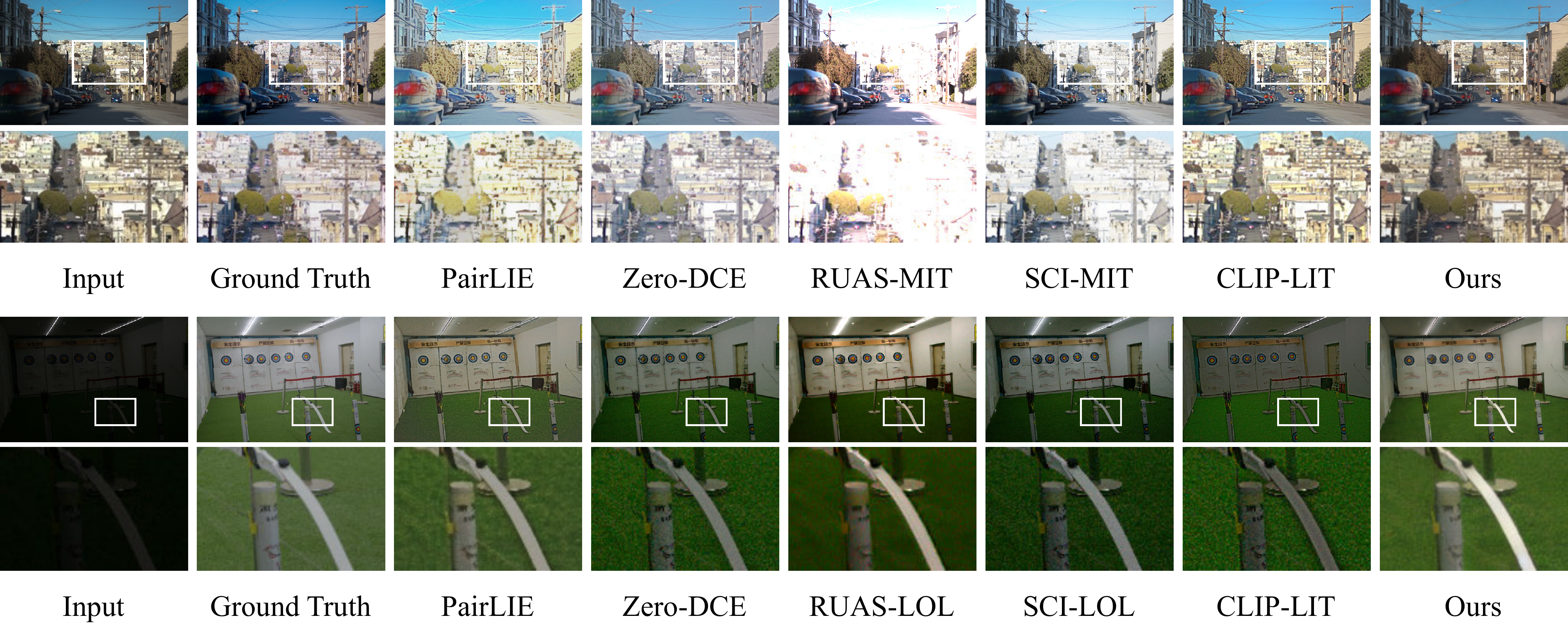}
    \vspace{-2mm}
    \caption{Example low-light enhancement results on the MIT-Adobe FiveK (top row) and LOL datasets (bottom row).}
    \label{fig:comp_lol}
\end{figure*}

\section{Experiments}

\subsection{Implementation Details}

\noindent \textbf{Framework Development.}
Our framework is trained on the COCO-2017~\cite{COCO} train and unlabeled set. We employ a minibatch size of 8, conducting training over 140k steps, approximately 5 epochs, with a learning rate set at 1e-4 and the ADAM optimizer \cite{kingma2014adam}.
To accommodate this sizable model within limited GPU memory, we implement float16 precision and DeepSpeed~\cite{DeepSpeed}.
More implementation details can be found in the supplementary.

\vspace{1mm}
\noindent \textbf{Compared Methods.}
Our model is compared with nine unsupervised low-light image enhancement methods. Among these, EnlightenGAN~\cite{Enhance_EnlightenGAN}, PairLIE~\cite{PairLIE}, NeRCo~\cite{NeRCo}, and CLIP-LIT~\cite{CLIP_LIT} utilize unpaired low-light-related data. The remaining five methods, ExCNet~\cite{ExCNet}, ZeroDCE~\cite{Enhance_ZeroDCE}, ZeroDCE++\cite{Enhance_ZeroDCE++}, RUAS\cite{RUAS}, and SCI~\cite{SCI}, are zero-reference.
Additionally, we present the results of six supervised methods to show the upper bound of our task.

\vspace{1mm}
\noindent \textbf{Benchmark Settings.}
We report performance on three sets of widely-used low-light datasets. The first two sets are LOL and MIT-Adobe FiveK~\cite{FiveK}.
For LOL, we adopt the official test sets of LOL v1~\cite{Enhance_RetinexNet} and LOL v2~\cite{Enhance_DRBN}, resulting in 115 low-/normal-light image pairs.
For MIT, we follow Retinexformer~\cite{Retinexformer} to split 500 pairs for testing.
Additionally, we gather low-light images from LIME~\cite{Enhance_LIME}, NPE~\cite{wang2013naturalness}, MEF~\cite{ma2015perceptual}, DICM~\cite{lee2013contrast}, and VV~\cite{vonikakis2018evaluation} that have no ground truth.
This set is simply called the ``unpaired set".
On LOL and MIT, we report PSNR, SSIM, LPIPS~\cite{zhang2018unreasonable}, and LOE \cite{wang2013naturalness}.
On the unpaired set, we report BRISQUE~\cite{mittal2012no} and noise level (NL) estimated by \cite{NoiseLevel}.

\subsection{Benchmarking Results}

Tab.~\ref{tab:lol_fivek_unpaired} demonstrates that our model surpasses the majority of unsupervised techniques and notably reduces the performance gap compared to supervised methods.
Subjective results can be found in Fig.~\ref{fig:comp_lol} and Fig.~\ref{fig:teaser}.
Our model is able to more effectively suppress noise and prevent overexposure or excessive darkness. More showcases can be found in the supplementary.
SCI~\cite{SCI} demonstrates good performance on training-related datasets but significantly degrades in unseen scenarios. In contrast, our model exhibits robustness across both LOL and MIT datasets simultaneously. This resilience stems from our model's ability to learn comprehensive illumination knowledge from the physical quadruple prior and normal light images. Consequently, our model proves to be less sensitive to specific datasets.

In Tab.~\ref{tab:lol_fivek_unpaired}, it's noticeable that supervised methods often tend to overfit to their training sets, exhibiting limited generalization to unseen domains.
For instance, Retinexformer~\cite{Retinexformer} and DiffLL~\cite{Diffll}, trained with LOL, achieve lower performance than our model on MIT, and vice versa.
This experiment underscores our model's superior adaptability to previously unseen scenarios, outperforming even supervised methods.

\subsection{Ablation Studies}
\label{sec:ablation}

\begin{figure*}[t]
    \centering
    \includegraphics[width=0.99\linewidth]{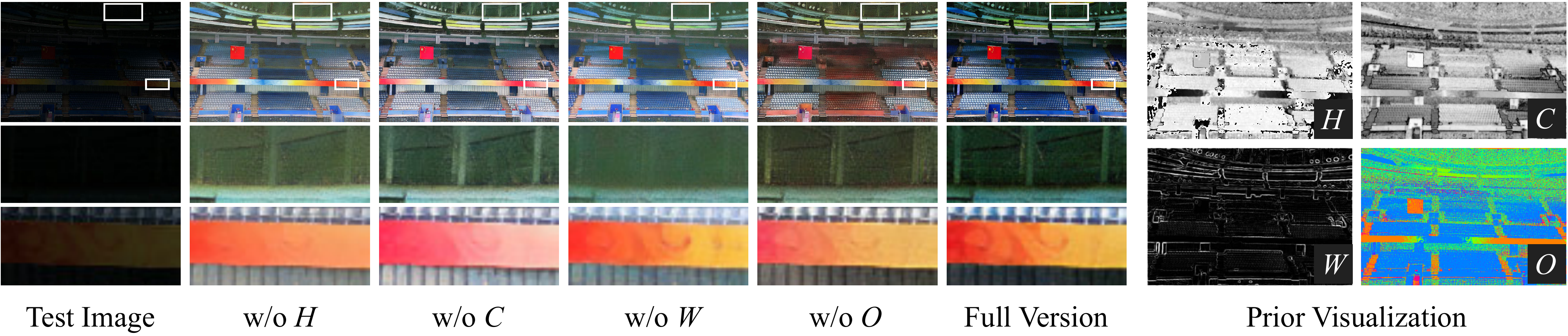}
    \vspace{-2mm}
    \caption{Low-light enhancement effects for different prior designs (left), and the visualization of our physical quadruple prior (right).}
    \vspace{-2mm}
    \label{fig:prior}
\end{figure*}

\vspace{1mm}
\noindent \textbf{Prior Design.}
We first analyze the effect of each element, $H$, $C$, $W$, and $O$, in our illumination-invariant prior.
The evaluation is conducted on LOL, following the same setting as Tab.~\ref{tab:lol_fivek_unpaired}.
Removing any element reduces the performance.
It is because deleting any element would remove a corresponding part of information.
Showing that only the four prior combinations together can reconstruct the original image without extracting the light-related features.

Fig.~\ref{fig:prior} provides a visual comparison.
The absence of $H$ or $C$ leads to color bias or a washed-out white appearance. As previously discussed in Sec.~\ref{sec:prior}, $H$ and $C$ are associated with hue and chroma.
But on the right of Fig.~\ref{fig:prior}, we can also see that $H$ and $C$ are not exactly hue and chroma.
This shows that prior basically conforms to the physical explanation we have derived, but can go on to learn more advanced features.
Recall that $W$ represents intensity-normalized spatial derivatives of spectral intensity, capturing local illumination changes. When $W$ is omitted, the alterations in light and shadow are completely lost (as observed in the second row of Fig.~\ref{fig:prior}). Additionally, without $O$, blue was mistakenly enhanced to orange.
Ultimately, our full version showcases the most refined details and superior contrast.

We further explore the implications of replacing our prior with alternative representations. We consider three representative ones:
(1) Naive HS channels in the HSV color space.
(2) CIConv~\cite{CIConv}, a trainable prior similar to our $W$.
(3) The reflectance estimated by Retinex-based PairLIE~\cite{PairLIE} trained on LOL.
In Tab.~\ref{tab:ablation_study}, the HS channels sacrifice significant content information, particularly impacting SSIM and LPIPS negatively.
While CIConv exhibits illumination invariance in high-level vision tasks, it suffers from excessive color loss in the image, leading to performance degradation.`
In comparison with our prior learned from COCO, the reflectance derived by PairLIE originates from low-light data pairs in LOL. Despite being trained on the target domain data, it still performs worse than our prior.

\begin{figure}
    \centering
    \includegraphics[width=0.99\linewidth]{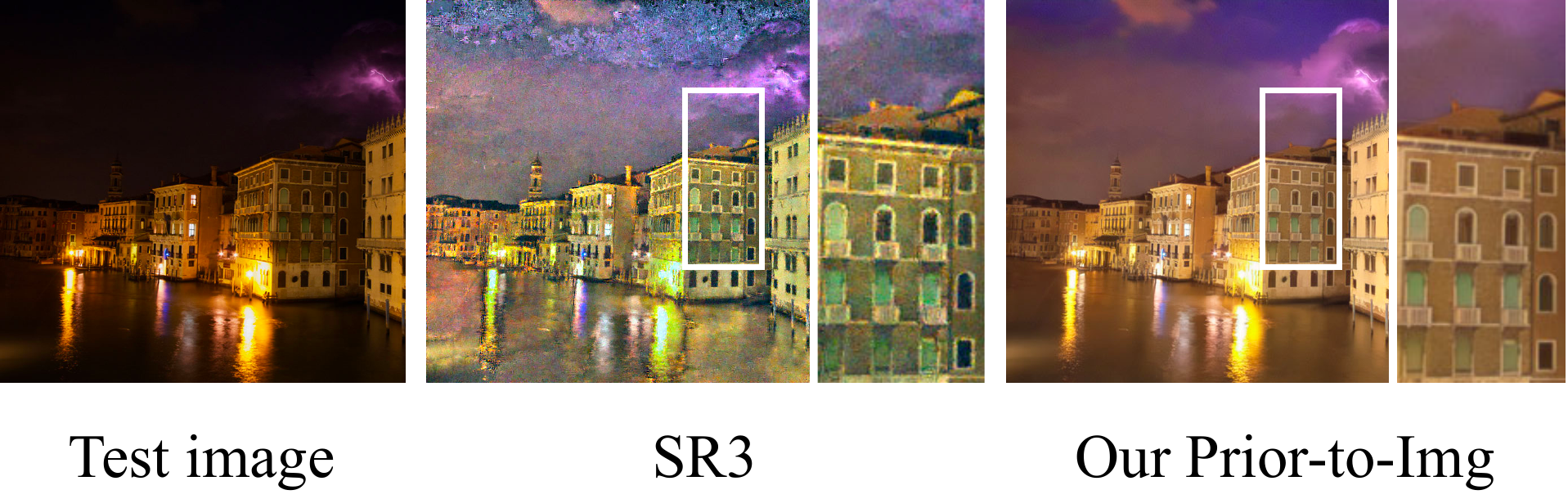}
    \vspace{-2mm}
    \caption{Effects of using different prior-to-image frameworks.}
    \label{fig:sr3}
\end{figure}

\vspace{1mm}
\noindent \textbf{Prior-to-Image Framework.}
We discuss the importance of using a pre-trained generative model to build the prior-to-image mapping.
We explore an alternative diffusion-based backbone, SR3~\cite{SR3}, recognized for its efficacy in super resolution.
However, when used to replace our prior-to-image framework, as illustrated in Fig.~\ref{fig:sr3}, SR3 exhibits noticeable color bias and noise issues.
It is because, in order to strip away light-related features, our prior discarded some of the image information, requiring the prior-to-image model to fill in the gaps and restore the complete details.

\begin{figure}
    \centering
    \includegraphics[width=0.99\linewidth]{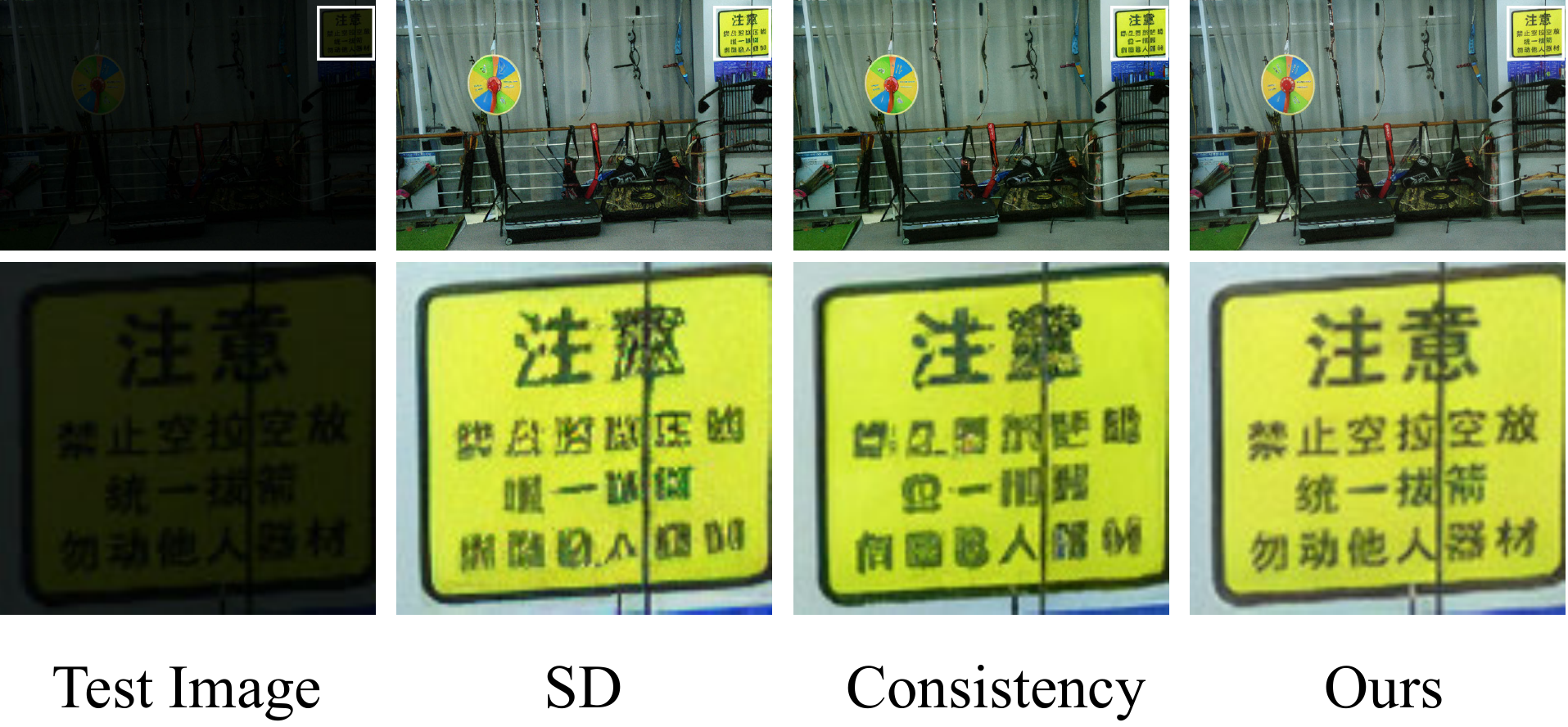}
    \vspace{-2mm}
    \caption{Effects of different decoders in our framework.}
    \label{fig:ae_lowlight}
\end{figure}

\begin{table}[t]
    \centering
    \caption{Ablation studies on the effect of our method designs.}
    \scriptsize
    \vspace{-2mm}
    \begin{tabular}{l|l|cccc}
        \toprule
        \multicolumn{2}{l|}{Datasets} & \multicolumn{4}{c}{LOL~\cite{Enhance_RetinexNet}} \\
        \midrule
        \multicolumn{2}{l|}{Metrics} & PSNR$\uparrow$ & SSIM$\uparrow$ & LPIPS$\downarrow$ & LOE$\downarrow$ \\
        \midrule			
        & Ours w/o $H$ & 17.60  & 0.756  & 0.262 & 0.314\\
        & Ours w/o $C$ & 17.60 & 0.762 & 0.262 & 0.313 \\
        & Ours w/o $W$ & 17.77 & 0.749 & 0.291 & 0.313 \\
        \multirow{2.5}{*}{Prior}  & Ours w/o $O$ & 18.63 & 0.764 & 0.285 & 0.315 \\
        \cmidrule{2-6}
        & HS channels in HSV & 18.04 &  0.562  &  0.498  & 0.410 \\
        & CIConv & 17.02 & 0.455 & 0.551 & 0.421\\
        & Reflectance by PairLIE~\cite{PairLIE} & 20.16 & 0.790 & 0.287 & 0.296\\
        \midrule
        \multirow{2}{*}{AE}& SD Decoder~\cite{LDM} & 19.26 & 0.665 & 0.243 & 0.353 \\
        & Consistency Decoder~\cite{DALLEE3} & 19.35 & 0.686 & 0.235 & 0.350\\
        \midrule
        \multicolumn{2}{l|}{Ours Final Version} & 20.25 & 0.807 & 0.199 & 0.278 \\
        \bottomrule
    \end{tabular}
    \label{tab:ablation_study}
\end{table}

\vspace{1mm}
\noindent \textbf{Auto-Encoder.}
We showcase the impact of our bypass decoder, comparing it with the original decoder used in SD. Additionally, we evaluate the Consistency Decoder from DALL-E 3~\cite{DALLEE3}, a recent diffusion-based decoder enhancing SD VAEs' decoding capabilities.
As shown in Fig.~\ref{fig:ae_lowlight}, both the original decoder in SD and the Consistency Decoder fail to preserve the text. In contrast, our decoder utilizes illumination-irrelevant details from input images, producing clear and undistorted text.
Furthermore, we evaluate the impact of the decoders in another image restoration task based on SD: colorization. Due to space limit, please refer to the supplementary for corresponding results.

\begin{figure}
    \centering
    \includegraphics[width=0.99\linewidth]{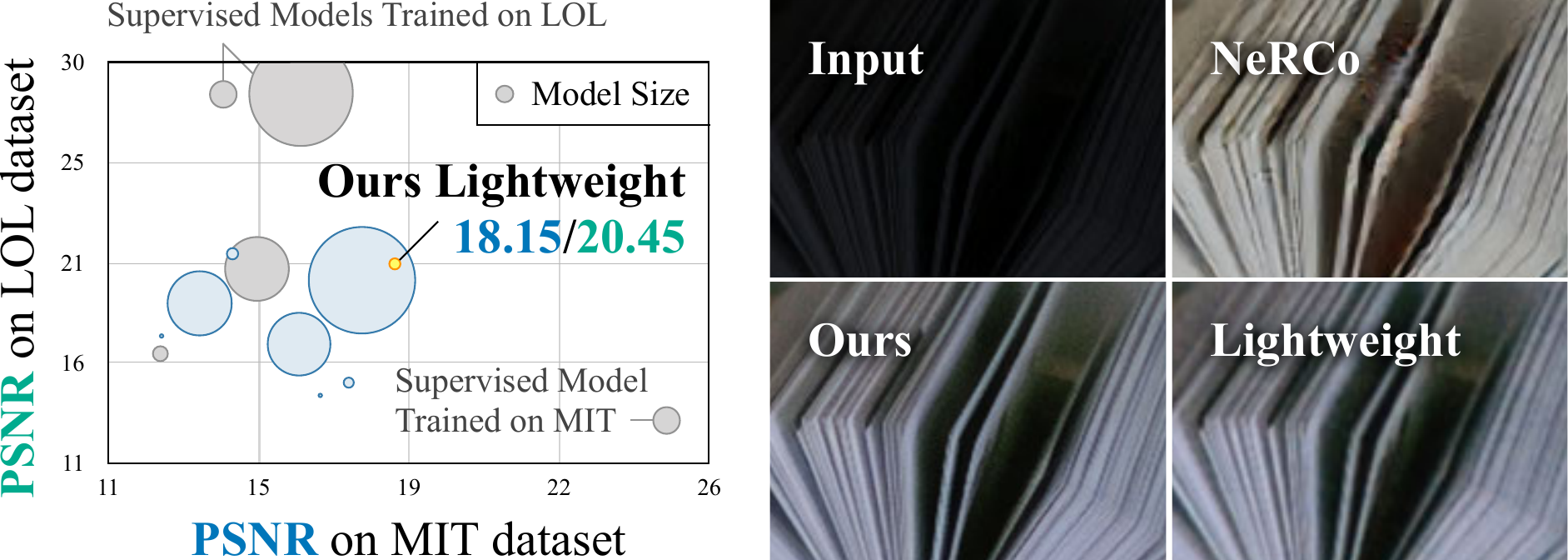}
    \caption{Left: Model size and PSNR comparison between our lightweight model and the SOTA. Detailed scores are in the supplementary. Right: Visual comparison of our full \& lightweight versions and NeRCo~\cite{NeRCo}. Both our full \& lightweight versions similarly show improved contrast and reduced color bias.}
    \label{fig:small}
    \vspace{-2mm}
\end{figure}

\vspace{1mm}
\noindent \textbf{Framework Distillation.}
Compared with the full model, our lightweight version reduces the running time to 500x faster.
It can process a 1024$\times$1024 image in 0.03 seconds on a Tesla M40. Additionally, the number of parameters is reduced from 1.3G to 327.36k, even smaller than SOTAs with comparable performance, as shown in Fig.~\ref{fig:small}.
The performance of our lightweight model on LOL/MIT datasets for PSNR, SSIM, LPIPS, and LOE is 20.45/18.15, 0.798/0.770, 0.290/0.175, and 0.273/0.174, respectively. 
Compared with our full model, the lightweight one achieves comparable performance and even marginally improves the PSNR and LOE.
This difference might be due to the randomness inherent in the generative diffusion model. During the fitting of the larger full version model, randomness is averaged, further reducing noise and rectifying errors.
\section{Conclusion}
We introduce a new zero-reference low-light enhancement framework, developed without low-light data. At its core lies a physical quadruple prior derived from the light transfer theory, and an efficient prior-to-image framework based on generative diffusion models. Experimental results show our superior performance across diverse scenarios.
{
    \small
    \vspace{-8mm}
    \bibliographystyle{ieeenat_fullname}
    \bibliography{main}
}

\end{document}